\documentclass[letterpaper, 10 pt, conference]{ieeeconf}  % Comment this line out if you need a4paper

\IEEEoverridecommandlockouts                              % This command is only needed if 
\usepackage{amsmath}
\usepackage{amssymb}             %math formula
\usepackage{amsfonts}            %math fonts
\usepackage{mathrsfs}            %math fonts

\usepackage{amsthm}
\usepackage{graphicx}
\usepackage{float}
\usepackage{booktabs}
\usepackage{multirow}
\usepackage{outline}
\usepackage{paralist}
\usepackage{siunitx}
\usepackage[table, dvipsnames, xcdraw]{xcolor}
\usepackage{subcaption}
\usepackage{adjustbox}
\usepackage{xspace}
\usepackage{balance}
\usepackage{hyperref}
\usepackage{pifont}
\usepackage{wrapfig}
\usepackage{tabularx}
\usepackage{makecell}
\usepackage{amsmath,amsfonts,bm}
\def\eqref#1{equation~\ref{#1}}
\def\1{\bm{1}}

\DeclareMathAlphabet{\mathsfit}{\encodingdefault}{\sfdefault}{m}{sl}
\SetMathAlphabet{\mathsfit}{bold}{\encodingdefault}{\sfdefault}{bx}{n}

\newcommand{\best}[1]{\textbf{#1}}         % handy macro

\definecolor{nhred}{RGB}{178, 34, 34}

\newcommand{\acronym}{HMC}

\definecolor{deepgreen}{RGB}{63, 126, 49}
\definecolor{deepred}{RGB}{196, 49, 25}
\definecolor{deepblue}{RGB}{0, 0, 139}

\theoremstyle{definition}

\definecolor{deepblue}{HTML}{27a2c3}
\definecolor{es-blue}{RGB}{0, 114, 178}
\usepackage
[sortcites,
backend=bibtex,
bibstyle=ieee,
citestyle=numeric,
sorting = nyt,
mincitenames=1,
maxcitenames=2,
doi=false,
isbn=false,
url=false,
eprint=false]{biblatex}

\makeatletter

\title{\bf \acronym{}: Learning Heterogeneous Meta-Control for  Contact-Rich Loco-Manipulation}

\author{
  Lai Wei\textsuperscript{*},
  Xuanbin Peng\textsuperscript{*},
  Ri-Zhao Qiu,
  Tianshu Huang,
  Xuxin Cheng,
  Xiaolong Wang 
    \\  $^{*}$equal contribution% <-this % stops a space
\vspace{0.15cm}
    \\  UC San Diego% <-this % stops a space
\vspace{0.1cm}
  \\ \texttt{\url{https://loco-hmc.github.io}}
}

\begin{document}
\twocolumn[{%
\renewcommand\twocolumn[1][]{#1}%
\maketitle
\begin{center}
    \centering
    \vspace{-18px}
    \captionsetup{type=figure}
    \includegraphics[width=\linewidth]{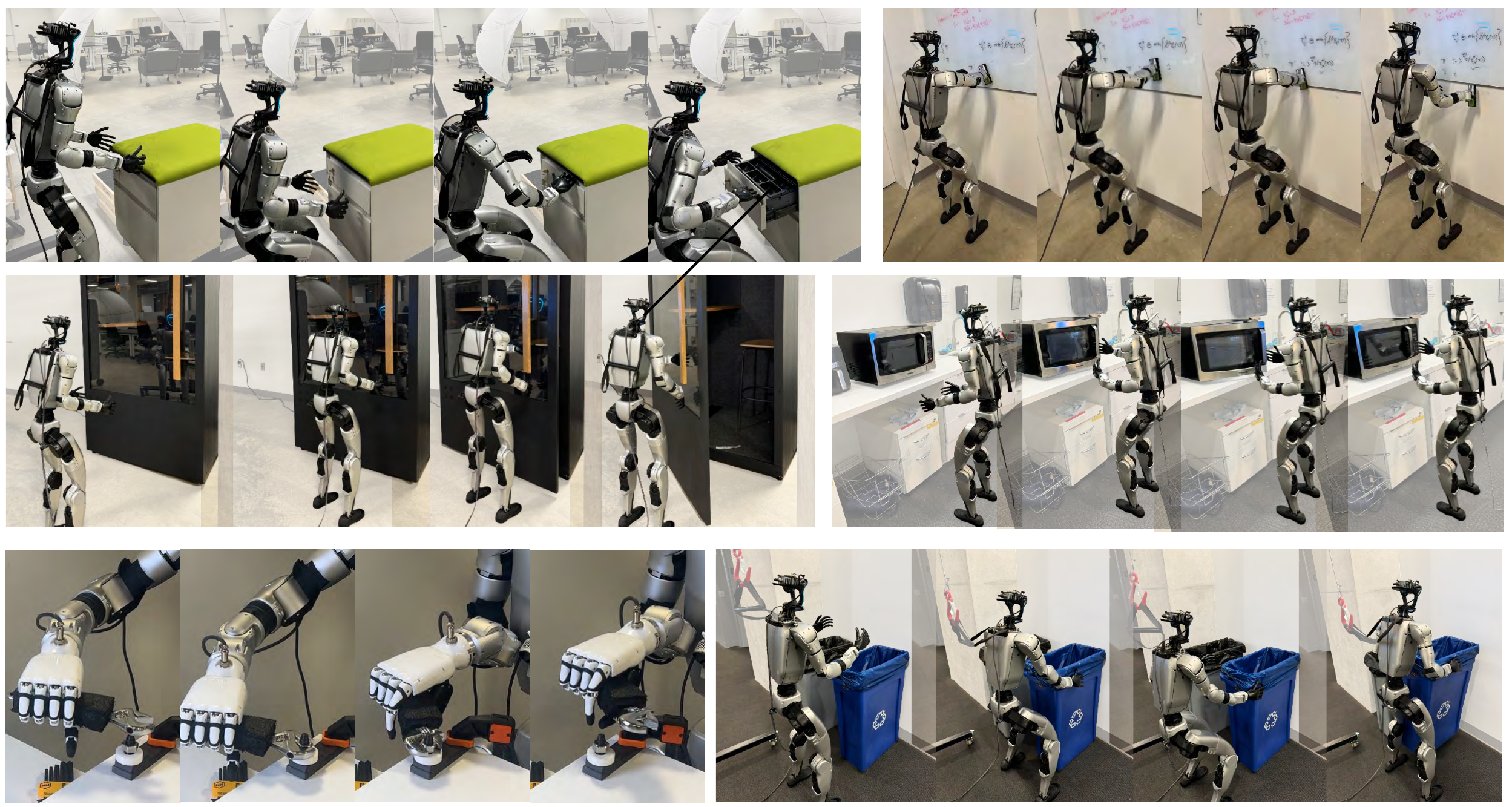}
    \caption{\textbf{Rolling out \acronym{} for contact-rich tasks on a humanoid robot.} Compared to na\"ive position-only policies~\cite{black2024-pi0,liu2024-rdt,kim2024-openvla,ghosh2024-octo}, \acronym{} designs a blending interface (HMC-Controller) that dynamically balances the torques from position tracking and force-aware controllers for contact-rich tasks. Using a heterogeneous architecture, \acronym{}-Policy trains on both large-scale position-only data and fine-grained force-aware demonstrations. (Shown tasks: opening a wheeled cabinet; wiping a whiteboard; pulling open a door; opening a microwave; tightening a nut and lifting a trash can.)}
    \label{fig:teaser}
\end{center}
}]

\begin{abstract}
Learning from real-world robot demonstrations holds promise for interacting with complex real-world environments. However, the complexity and variability of interaction dynamics often cause purely positional controllers to struggle with contacts or varying payloads. To address this, we propose a \textbf{H}eterogeneous \textbf{M}eta-\textbf{C}ontrol (HMC) framework for Loco-Manipulation that adaptively stitches multiple control modalities: position, impedance, and hybrid force-position. We first introduce an interface, \textbf{HMC-Controller}, for blending actions from different control profiles \textbf{continuously in the torque space}. HMC-Controller facilitates both teleoperation and policy deployment. Then, to learn a robust force-aware policy, we propose \textbf{HMC-Policy} to unify different controllers into a heterogeneous architecture. We adopt a mixture-of-experts style routing to learn from large-scale position-only data and fine-grained force-aware demonstrations. Experiments on a real humanoid robot show over 50\% relative improvement vs. baselines on challenging tasks such as compliant table wiping and drawer opening, demonstrating the efficacy of \acronym{}.
\end{abstract}

\vspace{-3px}

% \keywords{Loco-Manipulation, Contact-Rich, Heterogeneous Learning}

\section{Introduction}

Robots that walk and manipulate in a seamless whole --- known as \textit{loco-manipulation} systems --- extend robot autonomy beyond structured environments to enable versatile operations in homes, warehouses, and disaster sites. Recent advancements in large-scale learning methods, such as imitation learning~\cite{liu2024-rdt,black2024-pi0,2023-openxembodiment,kim2024-openvla,ghosh2024-octo}, sim-to-real transfer~\cite{lin2025-sim-humanoid,liu2024-vbc}, and zero-shot planning~\cite{ji2024-graspsplats,shen2023-F3RM,rashid2023-lerftogo,huang2024-rekep}, have driven significant progress in robot manipulation. Internet-scale pre-trained visual models ~\cite{liu2024-groundingdino,radford2021-clip,oquab2023-dinov2} and sophisticated teleoperation interfaces~\cite{zhao2023-ACT,cheng2024-opentv,iyer2024-openteach,chi2024-umigripper} have notably enhanced robotic capabilities.

However, these state-of-the-art methods primarily rely on \textit{position-only controllers}~\cite{liu2024-rdt,black2024-pi0,2023-openxembodiment,kim2024-openvla,ghosh2024-octo} that track only joint angles or end effector poses, which excel in tracking accuracy yet fundamentally neglect complex interaction dynamics. On the other hand, real-world tasks inherently involve \textbf{contact-rich interactions}, such as wiping surfaces, lifting heavy or delicate objects, or opening drawers secured by magnets. These scenarios demand precise yet adaptable regulation of both \textit{motion} and \textit{force}. Consequently, these controllers frequently generate hazardous oscillations and excessive forces in contact-rich tasks.

Traditional compliance methodologies, including impedance and admittance control~\cite{10.1115/1.3140702, doi:10.1177/027836498400300101,wbc-qm}, as well as hybrid force-position control~\cite{10.1115/1.3139652, 12073}, address some of these limitations by either dynamically adapting stiffness~\cite{10.1115/1.3140702, doi:10.1177/027836498400300101} or precisely managing forces in normal and shear directions~\cite{10.1115/1.3139652, 12073}. However, traditional approaches typically focus on a single, manually-tuned setting, lacking generalizability across different scenes. To address such an issue, recent methods~\cite{liu2025factrforceattendingcurriculumtraining,xue2025reactive,hou2024adaptive} incorporate learning-based strategies to learn force-aware actions in a data-driven fashion. FACTR~\cite{liu2025factrforceattendingcurriculumtraining} and RDP~\cite{xue2025reactive} use force feedback to handle contact-rich interactions, while ACP~\cite{hou2024adaptive} use imitation learning to dynamically adjust stiffness parameters. However, existing learning-based methods often rely on a specific type of compliant controller and a small set of domain-specific expert data collected by the exact controller, in addition to requiring expensive force sensors.

This paper identifies three types of challenges to develop a robust force-aware robot policy:
\begin{enumerate}
\item {\bf Modality Mismatch:} Position, impedance, and hybrid controllers each excel in different task phases, but no single mode is sufficient for the entire task.
\item {\bf Data Imbalance:} Large-scale teleoperation data is predominantly positional, whereas demonstrations leveraging compliance or force control are scarce.
\item {\bf Abrupt Switching:} Hard, discrete transitions between controllers cause torque discontinuities, leading to unstable and unsafe interactions.
\end{enumerate}

To overcome these challenges, we introduce \textit{Heterogeneous Meta-Control (HMC)}.
Our meta controller, HMC-Controller, is a meta-control~\cite{meta-control} interface that takes inputs from multiple low-level control modalities (position, impedance, hybrid force-position) and computes unified torque actions. HMC-Controller operates on-the-fly in the torque space, which enables continuous blending of different control profiles based on the evolving task progress and environment. We then build HMC-Policy, which learns all control profiles simultaneously in a heterogeneous fashion. HMC-Policy predicts actions from different control profiles, which are sent to the low-level HMC-Controllers. With a soft Mixture-of-Expert (MoE) routing design, HMC-Policy autonomously weights different controllers across the task execution horizon. Unlike traditional discrete switching strategies, our soft router ensures smooth and interpretable transitions between controllers, prevents expert collapse from imbalanced data distributions, and produces stable behaviors vital for real-world deployment. Furthermore, the heterogenous design intuitively enables a pre-train-fine-tune paradigm to train on open-sourced position-only demonstrations.

Empirical evaluations on challenging tasks show that HMC significantly outperforms conventional fixed or discrete-switching baselines. In sum, our contributions are:
\begin{itemize}
\item \textbf{Unified low-level control interface.} \acronym{}-Controller blends different control profiles continuously in the torque space for teleoperation and policy rollouts.
\item \textbf{Heterogeneous high-level policy.} \acronym{}-Policy uses heterogeneous learning and MoE-style soft routing, which trains on abundant positional data and fine-grained multi-expert demonstrations.
\item \textbf{Real-world evaluation.} Evaluations on various contact-rich tasks to show notable enhancements in stability, compliance, and adaptability in real-world scenarios.
\end{itemize}
\section{Related Work}
\begin{figure*}[h!]
  \centering
  \includegraphics[width=\textwidth]{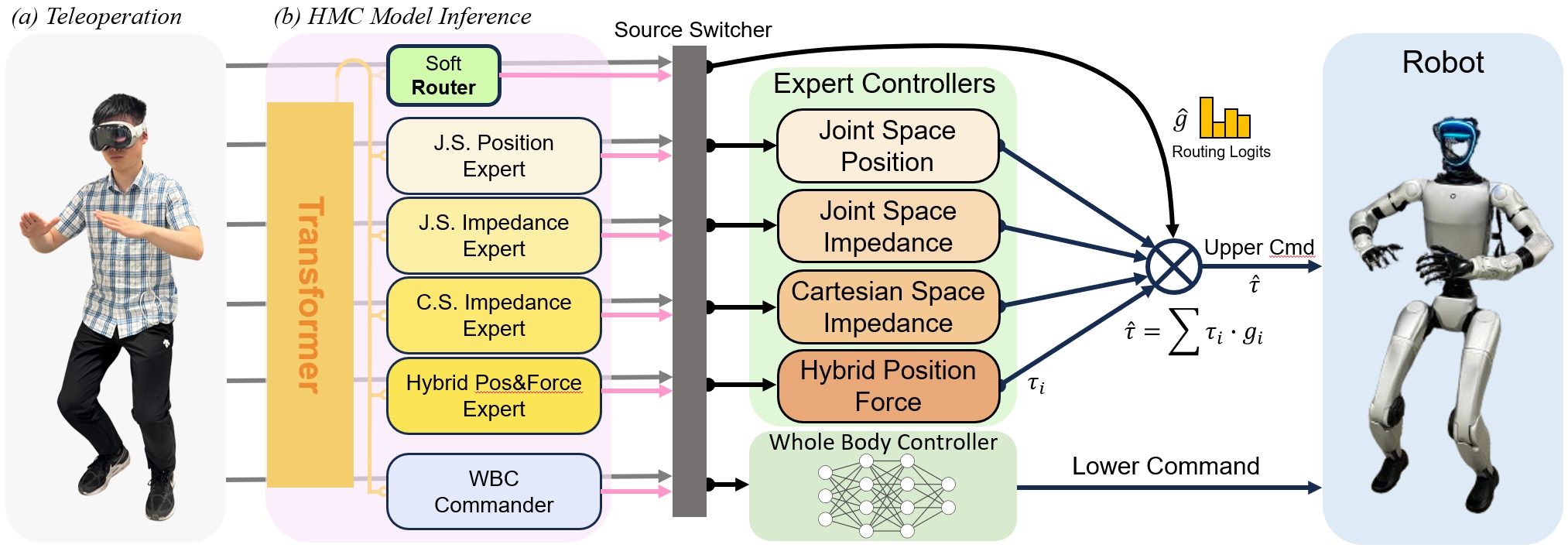}
  \caption{\textbf{System overview.} \acronym{}-Controller accepts inputs from either a VR-based teleoperation system or HMC-Policy inference. In the model inference path, multiple expert heads output corresponding control strategies. In the teleoperation path, shared hand poses are distributed to the expert controllers with different controller-specific parameters. All expert controllers output joint-space torque commands, which are blended via soft routing using predicted soft weights. Finally, the fused torque commands and lower-body joint targets are executed by the robot in real time. (``J.S'': Joint Space. ``C.S'': Cartesian Space.)}
  \label{fig:teleop_system}
\end{figure*}
\textbf{Manipulation Policies.}
Classical manipulation policies rely on model-based control or motion planning \cite{10.1145/3583136, zhou2021reviewmotionplanningalgorithms}, which often fail to generalize in unstructured real-world environments. In contrast, learning-based manipulation policies have seen rapid progress in recent years. Behavior Cloning \cite{bojarski2016endendlearningselfdriving} is a common approach to directly map from observations to actions. Recent developments include transformer-based models \cite{brohan2023rt1roboticstransformerrealworld, brohan2023rt2visionlanguageactionmodelstransfer, zhao2023-ACT} and diffusion-based methods \cite{chi2023-diffusionpolicy, zhang2024-diffusiondagger, ze20243ddiffusionpolicygeneralizable, janner2022planningdiffusionflexiblebehavior}, which aim to mitigate multi-modality and improve trajectory generation. Despite their success, these methods often rely on low-level positional controllers, making them ill-suited for contact-rich tasks that require reasoning about forceful interactions or multi-stage behaviors.

\textbf{Whole-body Loco-Manipulation.}
Loco-manipulation has emerged as an increasingly important direction in robotic control, especially in settings that demand both mobility and dexterous interaction. Early efforts such as \cite{fu2022deepwholebodycontrollearning, liu2024-vbc} primarily addressed static or quasi-static tasks, focusing on balance-constrained manipulation. More recent works including \cite{exbody2, gmt, ben2024homie, zhang2025falcon} tackle dynamic, contact-rich tasks that require whole-body coordination and physical stability during interaction. These systems highlight the growing demand for controllers that account for both task-specific objectives and the robot’s whole-body dynamics.

\textbf{Contact-Rich Compliance Control.}
Contact-rich manipulation often requires compliant behavior to maintain stability and prevent damage during interaction. Traditional model-based methods such as \cite{10.1115/1.3140702, 10.1115/1.3139652, doi:10.1177/0278364918768950} explicitly regulate force and motion based on physical models, but are difficult to tune and deploy in unstructured or dynamic environments. Learning-based approaches have gained traction, typically through reinforcement learning (RL) or learning from demonstration (LfD). Some RL-based methods \cite{Beltran_Hernandez_2020} predict compliance parameters in real time, which are then passed to a low-level controller. Others \cite{portela2024learning} integrate the RL agent directly into the low-level control loop. LfD methods \cite{hou2024adaptive, aburub2024diffusion, kamijo2024learning, liu2024forcemimic, wei2024ensuringforcesafetyvisionguided} often aim to infer both target motion and a set of compliance parameters from human demonstrations. Despite progress, these methods face a common challenge: collecting data with reliable force or impedance supervision remains difficult.

\textbf{Meta Controllers.}
% Emm……
Meta-controllers aim to automate the selection and coordination of robot skills, enabling agents to perform diverse tasks by dynamically invoking the most suitable experts. Prior approaches include skill library methods such as \cite{liang2024skilldiffuserinterpretablehierarchicalplanning, qiu2025wildlmalonghorizonlocomanipulation}, which retrieve or blend existing policies for new tasks. More recent efforts leverage large language models (LLMs) to guide skill selection, parameter inference, or reward specification, as seen in \cite{liang2023code, ma2023eureka}. Building on this line of work, \cite{meta-control} introduces a hierarchical control framework that synthesizes customized models and controllers per task via LLMs, mimicking human expert reasoning. However, these LLM-based methods often face latency constraints and struggle to integrate real-time onboard observations—such as contact feedback or phase transitions—into the control loop. This makes them ill-suited for multi-phase, contact-rich tasks that require fast and reactive expert switching. Our approach similarly adopts a structured hierarchy, but focuses on explicit phase switching and expert policy selection, enabling real-time, feedback-driven adaptation rather than offline synthesis.

\section{\acronym{}-Controller Interface}
% % 

Our loco-manipulation meta-control system is illustrated in Fig. \ref{fig:teleop_system}, which provides a unified interface for both teleoperation and deployment. We apply \acronym{} to the upper body of the robot for manipulation, with an off-the-shelf whole-body humanoid controller~\cite{li2025amo} to enable whole-body dexterity and enlarged workspace.

\textbf{HMC-Controller.} To get the best out of position-based controllers and force-aware controllers, we implement a suite of primitive controllers: pure position, impedance, and hybrid controllers. Each controller follows standard implementation and accepts desired Cartesian and/or joint setpoints along with a controller-specific profile. The outputs are torque commands for motors.

Given the robot joint positions $\mathbf{q}$, \textbf{pure position controller} is a standard PD controller~\cite{ang2005-pid} that directly tracks the desired joint positions $\mathbf{q}_d$ with gain coefficients $K_p$ and $D_p$.
\begin{equation}
\boldsymbol{\tau} = K_p (\mathbf{q}_d - \mathbf{q}) + D_p (\dot{\mathbf{q}}_d - \dot{\mathbf{q}})\,
\label{eq:position_pd}
\end{equation}
\textbf{Joint-space impedance controller} modulates compliance at the joint level with desired joint space stiffness $K_q$ and damping $D_q$.
\begin{equation}
    \boldsymbol{\tau} = K_q (\mathbf{q}_d - \mathbf{q}) + D_q (\dot{\mathbf{q}}_d - \dot{\mathbf{q}}) + \tau_{gravity}
    \label{eq:joint_impedance}
\end{equation}
\textbf{Cartesian-space impedance controller} regulates end-effector compliance in the Cartesian space using the Jacobian $J(\mathbf{q})$ and cartesian space stiffness $K_x$ and damping $D_x$:
\begin{equation}
    \boldsymbol{\tau} = J^\top \left[ K_x (\mathbf{x}_d - \mathbf{x}) + D_x (\dot{\mathbf{x}}_d - \dot{\mathbf{x}}) \right] + \tau_{gravity}
    \label{eq:cartesian_impedance}
\end{equation}
Finally, \textbf{hybrid position-force controller} tracks position along free axes while maintaining a specified force $\mathbf{f}_d$ along constrained directions, using selection matrices $S_p$ and $S_f$:

\begin{equation}
\boldsymbol{\tau} = J^\top \bigg[ 
     S_p \Big( K_x (\mathbf{x}_d - \mathbf{x}) 
    + D_x (\dot{\mathbf{x}}_d - \dot{\mathbf{x}}) \Big) \\
     + S_f (\mathbf{f}_d - \mathbf{f}) \bigg]  
    + \tau_{gravity}
\end{equation}
Given actions corresponding to different controllers at the same timestamp, \acronym{}-Controller computes torque for each controller, and performs a soft weighted average to obtain the final torques for execution. A low-pass filter is applied to ensure continuity and stability.

\textbf{Teleoperation Architecture.} We use OpenTV~\cite{cheng2024-opentv} to obtain head ($\mathbf{x}_{head}$) and hand ($\mathbf{x}_{hand}$) pose tracking at 50 Hz, which are solved to joint angles $\mathbf{q}_d$~\cite{caron2021pink,carpentier2019pinocchio}. In addition, our meta-control teleoperation system incorporates a user-friendly dashboard that enables the second operator to switch control modes and adjust parameters (e.g., impedance stiffness) on-the-fly. 

As a low-cost teleoperation solution, our system dispenses with dedicated physical control sticks (e.g., \cite{yang2024acecrossplatformvisualexoskeletonslowcost,liu2025factrforceattendingcurriculumtraining}) and end-effector force sensors. Instead, to convey cues during object interaction, we perform an online, coarse estimation of contact forces based on joint motor torques and positional errors. Together we visualize these forces on the virtual robot arm within the 3D scene to allow an immersive perception of contact dynamics than a conventional 2D display.

\begin{figure*}
  \centering
  \includegraphics[width=\linewidth]{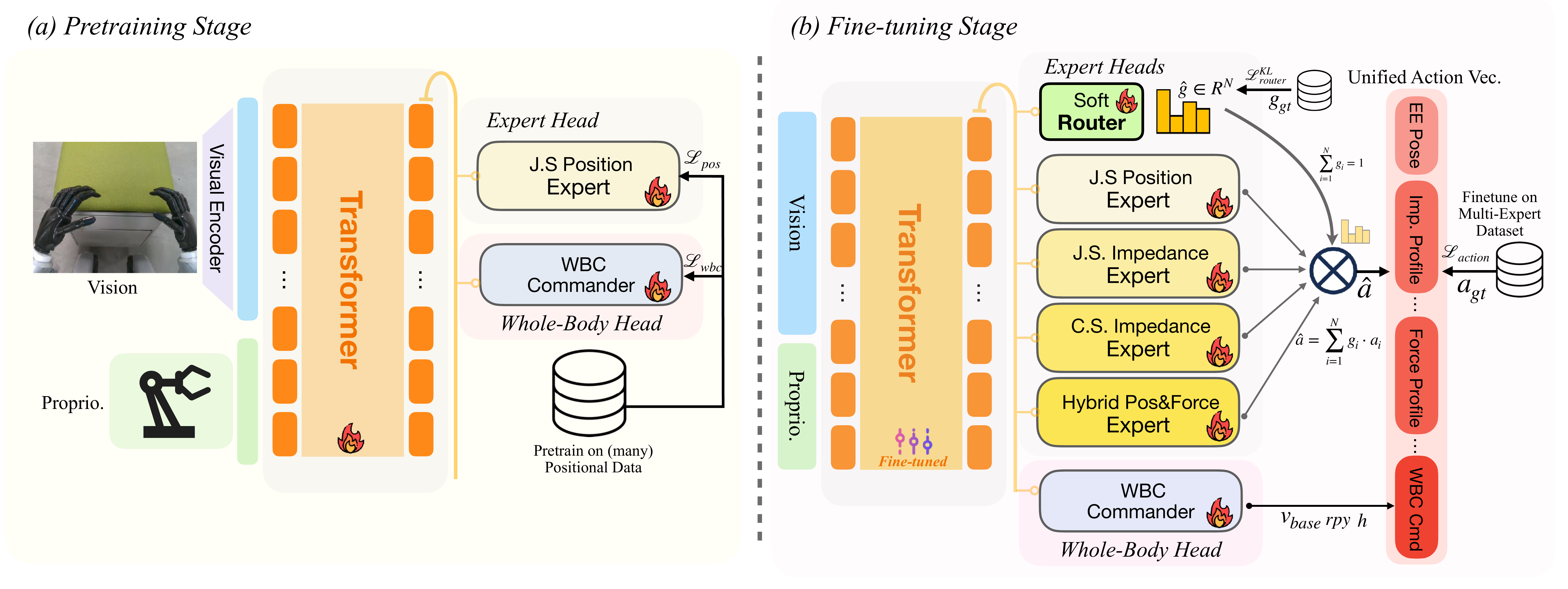}
  \caption{\textbf{Overview of Two-stage \acronym{}.}
\textbf{(a) Pretraining:} We harness abundant positional demonstrations to train the shared transformer trunk and position expert head, thereby embedding a strong positional prior that boosts generalization.  
\textbf{(b) Fine-tuning Stage:} All parameters are unfrozen and fine-tuned on a smaller, fine-grained multi-expert dataset. A soft routing network learns to blend outputs from multiple experts, producing smooth and adaptive control policies. (``J.S'': Joint Space. ``C.S'': Cartesian Space.)}
\vspace{-5px}
\end{figure*}
    \vspace{-5px}

\section{Learning Algorithm: Heterogeneous Meta-Control Policy}
\label{sec:algorithm}

\textbf{Problem Formulation and Motivation.} We frame the task of meta-control for loco-manipulation as a \textit{heterogeneous behavioral cloning} problem, where the goal is to replicate expert-level, multi-modal control behaviors demonstrated via teleoperation. Formally, we consider a demonstration dataset consisting of trajectories:
\begin{equation}
\tau = \{(o_t, a_t, g_t)\}_{t=1}^T, \quad a_t = \{\mathbf{EE}, \text{Imp. Stiff}, \cdots, \mathbf{F}\}
\end{equation}
where at each timestep $t$, the robot receives multi-sensory observations $o_t$, including visual $o_t^{\text{vis}}$ and proprioceptive $o_t^{\text{prop}}$ data. The action $a_t \in \mathbb{R}^{N_a}$ is unified across different controllers, including all potential values such as end-effector (EE) pose and attributes ({\it e.g.,} impedance stiffness, force vectors). For an action in the demonstration, zero-padding is applied for irrelevant controller attributes to maintain a consistent unified vector space. A soft control belief state $g_t \in [0,1]$ is used to indicate the relative contribution of each expert at time $t$, where $\sum_{i=1}^{N} g_{t,i}=1$.

Critically, teleoperation logs exhibit a highly imbalanced distribution—positional data dominate, while impedance and force modality data are scarce yet crucial for contact-rich interactions. A naive single-head network tends to collapse toward purely positional behavior, neglecting critical compliant behaviors. Conversely, separate specialist networks fail to handle smooth and continuous modality transitions required in real-world applications. To tackle these challenges, we propose a two-stage training strategy and a Soft Mixture-of-Experts (Soft MoE) architecture capable of dynamically blending control modalities smoothly.

\subsection{\acronym{}-Policy: Architecture and Two-Stage Training}

Our proposed Heterogeneous Meta-Control Policy (HMC-Policy) approach involves two main stages: (a) a positional pretraining stage leveraging abundant, easily acquired positional data, and (b) a fine-tuning stage utilizing fine-grained multi-expert demonstrations to enable smooth transitions among control modes. The architecture includes a shared Transformer-based encoder that extracts task-relevant latent embeddings, modality-specific expert heads, a soft gating mechanism (router), and a dedicated Whole-Body Control (WBC) commander head, which continuously outputs high-level whole-body directives (e.g., base velocity $V_{\text{base}}$, base height $h_{\text{base}}$, waist posture $rpy$, etc.) to the low-level HMC whole-body controller, ensuring stable and coordinated locomotion-manipulation behavior.

Positional data are cheap while contact dynamics vary widely; hence we adopt a shared Transformer trunk for general spatial reasoning and modality-specific experts that specialise in their respective dynamic regimes.

\textbf{Shared Transformer Trunk.}
Given the observation $o_t = (o_t^{\text{vis}}, o_t^{\text{prop}})$, we first tokenize each modality using separate tokenizers: a visual tokenizer $f_{\theta_{\text{vis}}}$ and a proprioceptive tokenizer $f_{\theta_{\text{prop}}}$. These tokenizers output fixed-length token embeddings:
\begin{equation}
v_t = f_{\theta_{\text{vis}}}(o_t^{\text{vis}}) \in \mathbb{R}^{N_v \times d}, \quad
p_t = f_{\theta_{\text{prop}}}(o_t^{\text{prop}}) \in \mathbb{R}^{N_p \times d},
\end{equation}
where $v_t$ and $p_t$ represent token sequences from visual and proprioceptive modalities respectively. The combined tokens are then passed into a Transformer trunk encoder $f_{\theta_{\text{trunk}}}$, which outputs a shared latent embedding $z_t = f_{\theta_{\text{trunk}}}(v_t, p_t) \in \mathbb{R}^{D}$.

\textbf{Modality-Specific Expert Heads.}
We construct $N$ modality-specific experts, each implemented as separate MLPs:
\begin{equation}
a_{t,i} = g_{\theta_i}(z_t), \quad i \in \{\text{pos}, \text{imp}, \text{hybrid force}\},
\end{equation}
where each expert outputs an action profile including the EE pose and modality-specific parameters (joint stiffness for joint-space impedance, Cartesian stiffness for Cartesian-space impedance, force vectors for hybrid pos\&force control, etc.).

\textbf{Soft Router.}
The router head, parameterized by $\phi$, predicts a gating weight (soft belief state) distribution:
\begin{equation}
g_t = \text{softmax}(r_{\phi}(z_t)), \quad g_t \in [0,1]^N, \quad \sum_{i=1}^{N} g_{t,i} = 1\,.
\end{equation}
Soft routing prevents expert collapse under imbalanced data by encouraging exploring, yields smooth transitions between control modalities, avoids abrupt torque discontinuities and enhances hardware safety during deployment. The final unified action prediction at each timestep is computed by blending the expert outputs weighted by the router outputs $\hat{a}_t = \sum_{i=1}^{N} g_{t,i} \cdot a_{t,i}$.

\textbf{Whole-Body Control (WBC) Commander.}
A WBC commander head, $h_{\psi}(z_t)$, continuously outputs whole-body high-level commands, which are concatenated into the final unified action vector alongside the blended expert predictions.
\begin{equation}
a_{t}^{\text{WBC}} = h_{\psi}(z_t).
\end{equation}

\textbf{Pretraining Stage (Positional Prior Learning).} Initially, we freeze all expert heads except the positional expert and WBC commander, and pretrain the Transformer trunk on large positional-only datasets. Positional data is abundant and easier to acquire, naturally covering diverse variations such as initial poses, recovery behaviors, and positional distributions. This embeds a strong positional prior into the shared representation:
\begin{equation}
\mathcal{L}_{\text{pretrain}} = \|a_t^{\text{pos}} - a_t^{\text{pos, gt}}\|_1 + \|a_t^{\text{WBC}} - a_t^{\text{WBC, gt}}\|_1\,.
\end{equation}
\textbf{Fine-tuning Stage (Multi-Expert Integration).} In the fine-tuning stage, all heads are unfrozen and co-trained, while the Transformer trunk receives a reduced learning rate. A balanced multi-expert dataset and the soft router keep gradients flowing to every expert, avoiding modality collapse and enabling smooth blending. The final fine-tuning objective is given by
% \textbf{Unified action reconstruction:}
\begin{equation}
\mathcal{L}_{\text{finetune}} = \lambda_{\text{action}} \cdot \underbrace{\|\hat{a}_t - a_t^{gt}\|_1}_{\mathcal{L}_{\text{action}}} + \lambda_{\text{router}} \cdot \underbrace{D_{\text{KL}}(g_t^{gt}\| g_t)}_{\mathcal{L}_{\text{router}}}\,,
\end{equation}
where $g_t^{gt}$ in $\mathcal{L}_{\text{router}}$ is a smoothed ground truth signal obtained via low-pass filtering of tele-operated modality IDs, and $\lambda$ are loss-balancing hyper-parameters. 

\section{Experiments}

% We structure our experiments to address the following key research questions:

\subsection{Experiment Setup}

\hfill

\begin{figure}[ht]
  \centering
  \begin{subfigure}[b]{0.156\textwidth}
    \includegraphics[width=\linewidth]{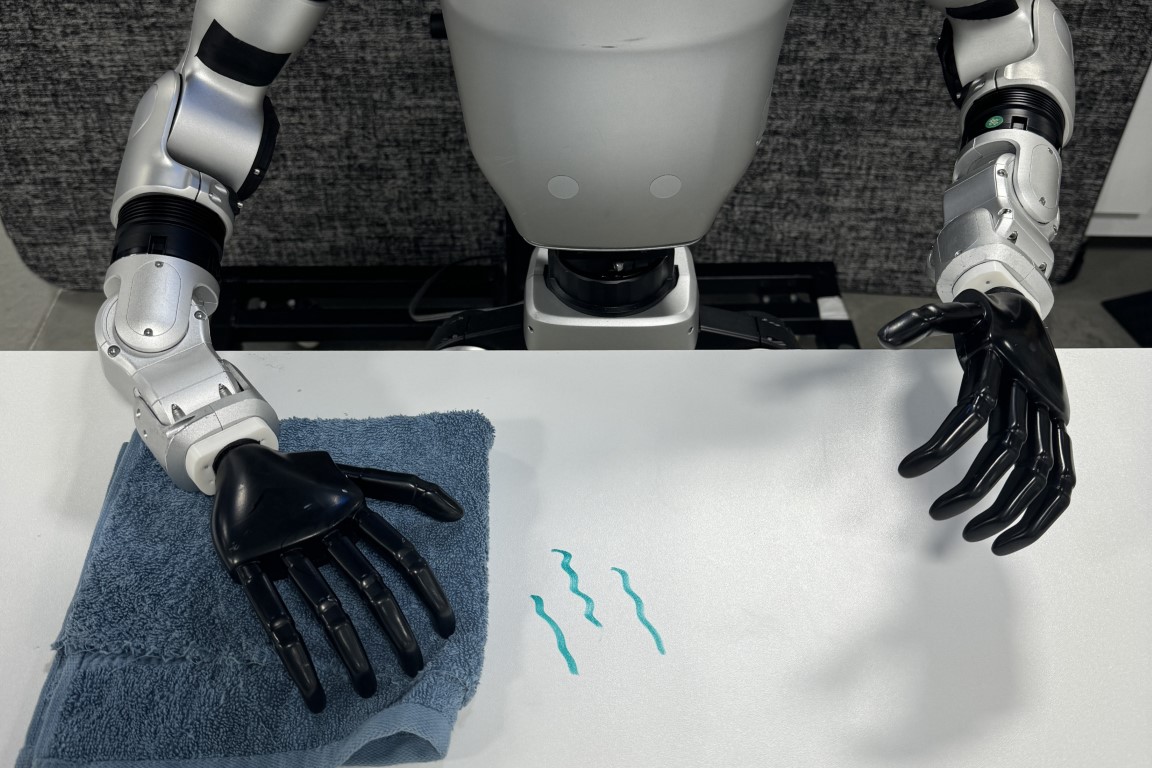}
    \caption{}
    \label{fig:wipetable}
  \end{subfigure}
  \hfill
  \begin{subfigure}[b]{0.156\textwidth}
    \includegraphics[width=\linewidth]{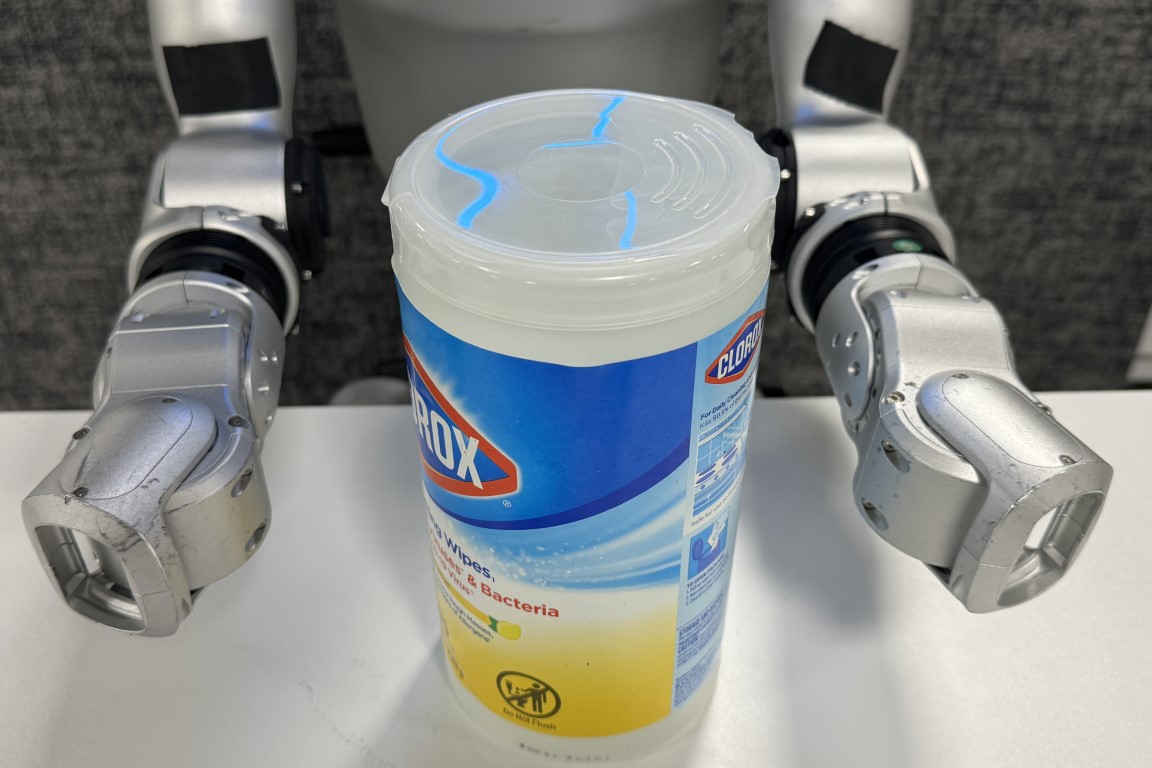}
    \caption{}
    \label{fig:lifebottle}
  \end{subfigure}
  \hfill
  \begin{subfigure}[b]{0.156\textwidth}
    \includegraphics[width=\linewidth]{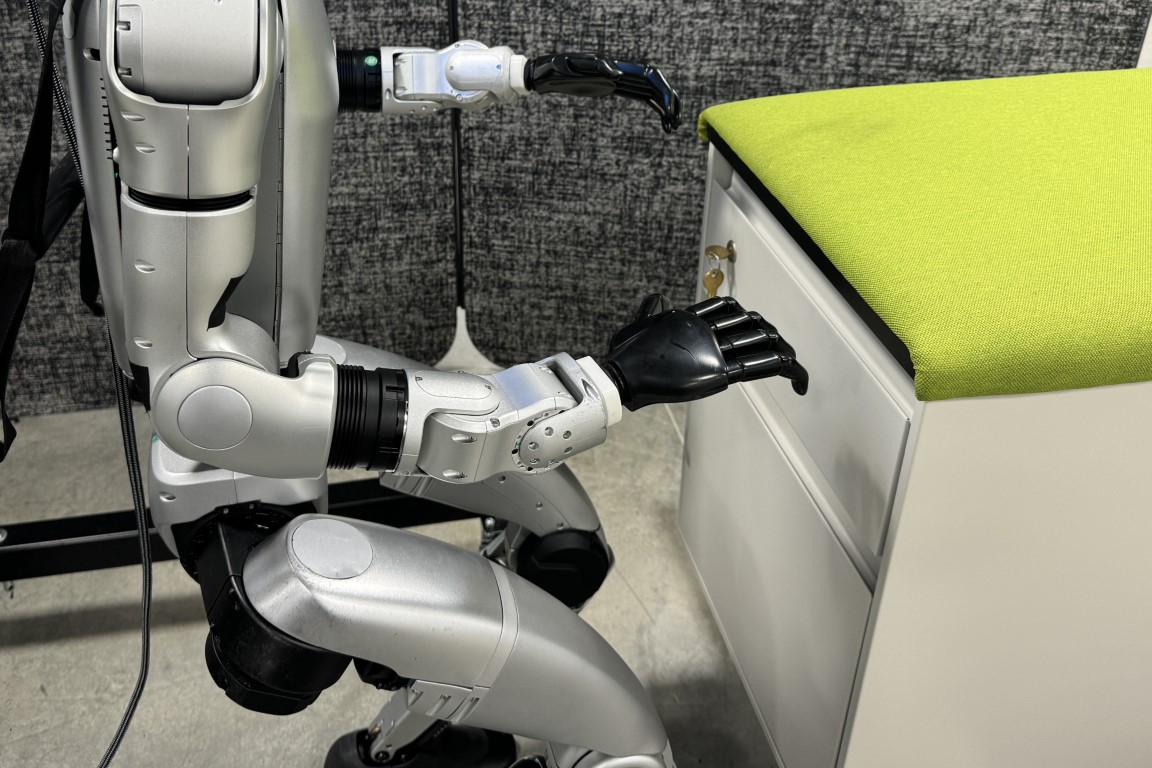}
    \caption{}
    \label{fig:opendrawer}
  \end{subfigure}
  \caption{Tasks visualization.}
  \label{fig:tasks}
\end{figure}

% ---------------- Table ----------------
\begin{table*}[t]
\centering\small
\begin{tabularx}{\textwidth}{l *{2}{>{\centering\arraybackslash}X} *{4}{>{\centering\arraybackslash}X}}
\toprule
\multirow{3}{*}{\textbf{Method}} &
    \multicolumn{2}{c}{\textbf{Lift Bottle(\%)}} &
    \multicolumn{4}{c}{\textbf{Open Drawer(\%)}} \\[2pt]
\cmidrule(lr){2-3}\cmidrule(lr){4-7}
 & Seen & Unseen &
   \multicolumn{2}{c}{Seen} & \multicolumn{2}{c}{Unseen} \\
 &  &  & \makecell{Hand Insert} & Pull &
        \makecell{Hand Insert} & Pull \\
\midrule
Stiff Policy            & 67  & 33  & 80  & 80  & 33  & 20  \\
Compliant Policy        & 80  & 40  & 80  & 53  & {67 } & 13  \\
HMC (w/o soft routing)  & {93 } & 53  & 80  & 80  & 53  & 47  \\
HMC (from scratch)      & 87  & 60  & 60  & 60  & 47  & 33  \\
\rowcolor{Gray!15}
HMC (ours)              & \best{93 } & \best{80 } &
                          \best{87 } & \best{87 } &
                          \best{67 } & \best{67 } \\
\bottomrule
\end{tabularx}
\caption{Ablation experiment results across seen and unseen settings for the Lift Bottle and Open Drawer tasks. The proposed soft routing and position-only pretraining increases success rate considerably. Hand insertion and pulling correspond to two separate stages in opening a drawer.}
\label{tab:ablation-result}
\vspace{-6pt}
\end{table*}

\begin{figure*}
  \centering
  \includegraphics[width=\linewidth]{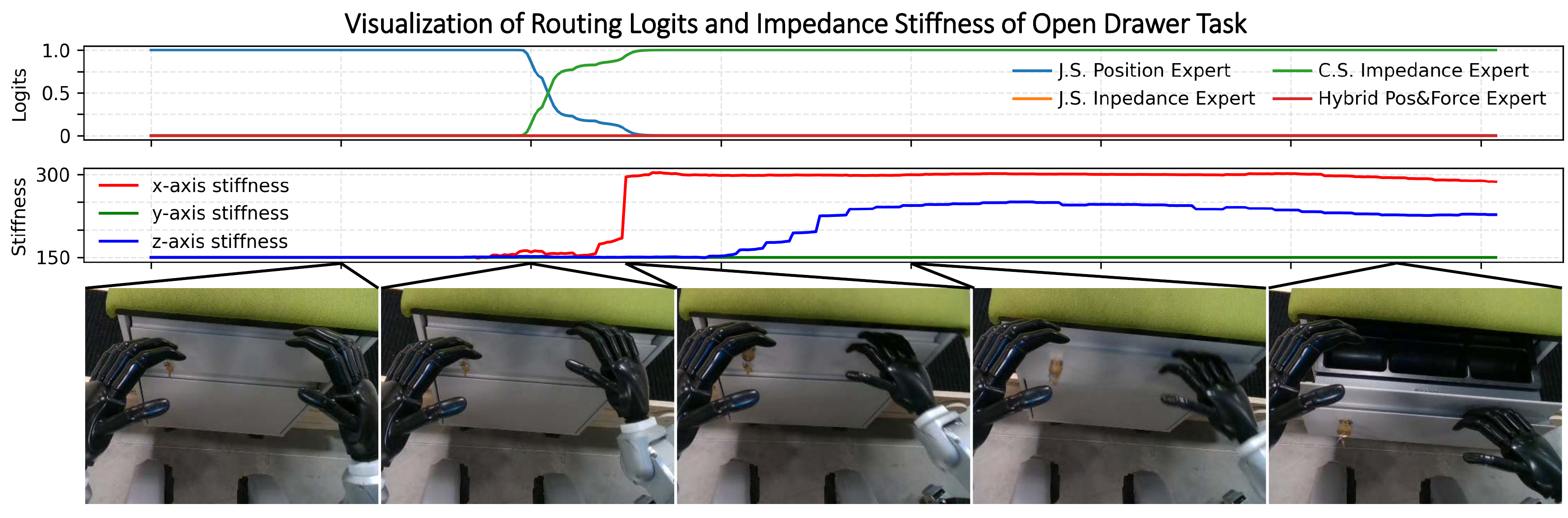}
  \caption{\textbf{Interpretability of routing logits} in visualization of an Open Drawer Episode. The upper figure is the predicted routing logits, while the lower figure shows the predicted right-hand stiffness ($N/m$). Five images from the head camera are demonstrated.}
  \label{fig:stagewise}
  \vspace{-5px}
\end{figure*}

\textbf{Hardware.}
We conduct experiments using the Unitree G1 humanoid robot, equipped with two 7-DoF arms for loco-manipulation tasks. 
The visual information is acquired from an Intel RealSense D435i camera mounted on the robot’s head, and all proprioceptive data are obtained via the Unitree SDK. 
The robot is controlled through our HMC-Controller interface (shown in Fig.~\ref{fig:teleop_system}), enabling seamless demonstration collection and precise multi-modal control execution.

\textbf{Task Setup.}
We evaluate our approach across three contact-rich loco-manipulation tasks (Fig.~\ref{fig:tasks}).
\textbf{Wipe Table:} the robot wipes the table surface to remove marker traces. This task requires \textbf{appropriate force regulation} capability. %—insufficient force fails to erase the marks, while excessive force may cause large unsafe movement.% A success is recorded if marks are wiped in two loops.
\textbf{Lift Bottle with Both Hands:}
The robot is tasked with lifting a bottle using only the bare end-effectors without grippers or hand, only relying on arm friction to secure the object. This task requires both bimanual coordination to maintain closed-chain dynamic contact and lift the bottle stably Precision. %, as the robot needs to lift bottles with different size, shape, and material. The task is considered successful if the bottle is lifted 10 cm above the table and held steadily for 3 seconds.
% The robot lifts a bottle using only the bare end-effector without a hand, relying on arm contact to secure the object. The task is successful if the bottle is raised 10 cm above the table and held stably for 3 seconds.
\textbf{Open Drawer:}
The humanoid crouches and inserts its hand into a narrow handle slot (Hand Insertion), then pulls to open a drawer secured by a magnetic latch, requiring over 20N of force to overcome closure resistance (Pulling). This task requires: (1) \textbf{stage-aware control} to switch between compliant insertion and forceful pulling stages to achieve high success rate; (2) \textbf{dynamic stiffness modulation} to adjust impedance properties during interaction with the drawer; and (3) \textbf{whole-body stability} to maintain lower-body balance while the upper body applies large interaction forces.% The task is considered successful if the drawer is pulled open by at least 20 cm.

\textbf{Baselines.} we implement the baselines as follows:
\begin{itemize}
\item \textbf{ACT (vanilla)}: Action Chunking Transformer (ACT) policy, trained solely on positional data.
\item \textbf{ACT (meta)}: ACT policy trained with positional data and controller profiles. Hard routing switch.
\item \textbf{Stiff Policy}: Our model architecture trained exclusively on stiff positional expert data.
\item \textbf{Compliant Policy}: Our model architecture trained exclusively on impedance experts data.
\item \textbf{HMC w/o soft routing}: Our HMC architecture with a hard \texttt{argmax} router instead of soft routing.
\item \textbf{HMC (from scratch)}: Our HMC architecture trained from scratch instead of two stages training.
\item \textbf{HMC (ours)}: The complete implementation of our proposed HMC-Policy.
\end{itemize}

\textbf{Evaluation Protocols.}
For the Open Drawer and Lift Bottle tasks, we evaluate generalization under seen and unseen settings, where the seen settings correspond to configurations encountered during training, and unseen ones involve novel object instances or initial positions for Lift Bottle and Open Drawer tasks respectively.
Each policy variant is evaluated over 15 trials, with performance assessed based on the task success rate. 
% Results across all variants and tasks are recorded, enabling detailed comparisons and robust evaluation of our proposed approach's efficacy and generalization capabilities.

\subsection{Evaluation Results}

The general experiment results across different policies are shown in Table \ref{tab:general-result}, and the ablation studies result are demonstrated in Table \ref{tab:ablation-result}. We try to answer important questions in our evaluation:

\begin{itemize}
    \item \textbf{Q1:} What advantages does our HMC-Policy have compared to baselines?
    \item \textbf{Q2:} Why do we need HMC-Controller for contact-rich and multi-stage tasks?
    \item \textbf{Q3:} What are benefits of Soft Routing? 
    \item \textbf{Q4:} Does pretrain and finetune approach helps training and generalization performance?
    \item \textbf{Q5:} Why do our HMC-Policy excels at contact-rich unseen tasks?
\end{itemize}

% \textbf{Q0: Does our teleoperation system helps to handle contact-rich multi-stage loco-manipulation tasks?}

\begin{table}[t]
\centering
\small
\begin{tabular}{lccc}
\toprule
 & Wipe (\%) & Lift (\%) & Drawer (\%) \\ \midrule
Stiff Policy           & 33 & 67 & 80 \\
ACT (vanilla)          & 33 & 60 & 40 \\
ACT (meta)             & 47 & 67 & 47 \\
HMC (w/o soft routing) & 87 & 93 & 80 \\
\rowcolor{Gray!15}
HMC (ours)             & \best{93} & \best{93} & \best{87} \\
\bottomrule
\end{tabular}
\caption{Policy success rates on three contact-rich tasks, including wiping table, lifting bottle and opening drawer.}
\label{tab:general-result}
\vspace{-5px}
\end{table}

\textbf{A1: Our HMC consistently outperforms baseline approaches across all three tasks.} 
The stiff policies, though widely used, struggle in contact-rich scenarios—for example, in Wipe Table their rigidity produced excessive torques, sometimes triggering motor overheating. ACT (meta) improves upon its vanilla version by jointly predicting trajectories and meta-controller profiles, but its single-head design makes joint learning of low-level commands and high-level modes difficult, often causing unstable behaviors in complex tasks. By contrast, HMC separates meta-level decision-making from trajectory generation and dynamically routes to expert controllers, reducing learning complexity and ensuring robust execution in multi-stage manipulation.

% \textbf{Q2: Why do we need meta controller for contact-rich and multi-stage tasks?} % (self ablation, compared to stiff and pure complaint)

\textbf{A2: HMC-Controller is essential as it enables adaptive control switching across task phases.} 
Contact-rich tasks require distinct strategies at different stages. In the Open Drawer example, insertion benefits from compliance to tolerate alignment errors, while pulling demands stiffness to overcome resistance. As shown in Table~\ref{tab:ablation-result}, fixed controllers (stiff-only or compliant-only) fail in unseen settings since neither adapts to both needs. By contrast, HMC-Policy flexibly routes between expert controllers (Fig.~\ref{fig:stagewise}), using compliance for insertion and increasing stiffness for pulling, thereby achieving higher success through phase-aware control routing.

% \textbf{Q3: What are benefits of Soft Routing? } % (Self ablation Hard Routing)
\textbf{A3: Soft routing enables smooth and stable transitions between expert controllers.}
This is critical in contact-rich tasks, where abrupt switching can destabilize the interaction. In HMC-Controller framework, soft routing allows for continuous blending of control outputs, ensuring robustness during phase transitions.
In the Lift Bottle task under unseen settings, we observed that the HMC-Policy without soft routing often triggered an abrupt expert controller switch when the arm made contact with the bottle. Subtle discrepancies between expert controllers caused sudden changes in gripping force, occasionally leading to slippage of the cylindrical bottle. By contrast, soft routing allowed the robot to interpolate control outputs across experts, maintaining stability during transitions.

% \textbf{Q4: Does pretrain and finetune approach helps training and genearlization performance?}

\textbf{A4: Pretraining followed by finetuning improves both training stability and generalization.}
Compared to the training-from-scratch variant in Table~\ref{tab:ablation-result}, the pretrain-and-finetune strategy significantly enhances performance, particularly in unseen scenarios. During pretraining on position-only data, the shared stem network learns fundamental task representations and spatial alignment. Subsequent finetuning on heterogeneous controller data focuses learning on expert controller outputs, leveraging a well-initialized stem. This two-stage process mitigates conflicts among expert losses and stabilizes gradients during training. Moreover, the pretrained stem provides a strong prior, improving generalization and reducing the risk of overfitting to training environments.

% \textbf{Q5: Why do our meta controller excels at contact-rich unseen tasks? }

\textbf{A5: Our HMC-Policy excels in contact-rich unseen tasks by combining adaptive control routing with strong generalization.}
Contact-rich tasks often involve unpredictable variations in contact dynamics, requiring adaptive and robust control strategies. Our HMC-Policy framework addresses this by combining meta-level scheduling of diverse expert controllers with smooth transitions enabled by soft routing. Additionally, the pretrain-and-finetune paradigm provides a strong initialization, improving the generalizability with existing position-only data.

\section{Conclusion}

Contact-rich loco-manipulation exposes the limits of single-mode, position-only control. In this work, we tackled this gap with Heterogeneous Meta-Control (HMC), a unified framework in which HMC-Controller blends heterogeneous commands directly in torque space, while HMC-Policy learns via a two-stage pretrain-finetune paradigm, and Mixture-of-Experts to route those modalities smoothly in response to task phase and contact dynamics. This heterogeneous design compensates for the heavy positional bias in teleoperation data, keeps all experts trainable, and removes the torque discontinuities that plague hard switching .

% Comprehensive real-robot experiments on the Unitree G1 across diverse tasks—table wiping, bimanual bottle lifting, and drawer opening—show that HMC-Policy consistently outperforms stiff, compliant-only, and hard-switching baselines, delivering higher success rates, smoother force profiles, and better generalization to unseen objects and configurations. Taken together, these results demonstrate that adaptive, soft routing of complementary control modalities is a practical path toward robust whole-body loco-manipulation in the wild. Future work will explore scaling to a broader array of experts (e.g., tactile-conditioned manipulation) and integrating high-level task planners to further extend long-horizon autonomy.
% Short version of paragraph above
Real-robot experiments on the Unitree G1 (wiping, bottle lifting, drawer opening) show that HMC-Policy surpasses various baselines, yielding higher success, smoother forces, and stronger generalization. These results highlight adaptive soft routing as a practical approach to robust loco-manipulation, with future work aimed at scaling to more experts and integrating task-level planners for long-horizon autonomy.
%===============================================================================

\label{sec:conclusion}

%===============================================================================

% \clearpage
\printbibliography

\end{document}